\documentclass[11pt]{article}

\usepackage[english]{babel}

\usepackage[letterpaper,top=2cm,bottom=2cm,left=3cm,right=3cm,marginparwidth=1.75cm]{geometry}

\usepackage{float}
\usepackage{multicol}
\usepackage{mathtools}
\usepackage{subfigure}
\usepackage{listings}
\usepackage{setspace}
\usepackage[UKenglish]{isodate}
\usepackage{amssymb,amsmath} 
\usepackage{graphicx}
\usepackage[colorlinks=true, allcolors=blue]{hyperref}

\newcommand{\ex}[1]{\mathbb{E}\left[#1\right]}
\newcommand{\exc}[2]{\mathbb{E}_{#1}\left[#2\right]}
\newcommand{\exn}[1]{\hat{\mathbb{E}}_n\left[#1\right]}

\newcommand{\dn}[0]{\mathcal{D}_n}
\newcommand{\p}[0]{\mathbb{P}}

\title{Grafting: Making Random Forests Consistent}
\author{Nicholas Waltz}

\begin{document}
\setlength{\parindent}{0pt}
\maketitle

\begin{abstract}
Despite their performance and widespread use, little is known about the theory of Random Forests. A major unanswered question is whether, or when, the Random Forest algorithm is consistent. The literature explores various variants of the classic Random Forest algorithm to address this question and known short-comings of the method. This paper is a contribution to this literature. Specifically, the suitability of grafting consistent estimators onto a shallow CART is explored. It is shown that this approach has a consistency guarantee and performs well in empirical settings.
\end{abstract}
\pagebreak
\section{Introduction}

This paper studies a variant of the canonical Random Forest \cite{breiman_random_2001} that addresses some of the short-comings commonly associated with the classical method. Specifically, its inconsistency in some settings and its inability to deal with high-frequency periodic patterns. It is shown that this variant has a consistency guarantee and performs well in empirical studies.\\

Random Forests are a non-parametric ensemble learning method that work by averaging a large number of tree learners. The standard approach due to Breiman \cite{breiman_random_2001} uses CARTs (Classification and Regression Trees) \cite{breiman_classification_1984}. They are known for their combination of fast computation time, interpretability and good performance across a wide set of applications \cite{howard_two_2012}. In particular, due to the underlying feature selection in the CART algorithm, they are particularly suited to high-dimensional settings \cite{scornet_consistency_2015, chernozhukov_doubledebiased_2018, klusowski_sparse_2020}, theoretically even when the ambient dimension exceeds the sample size. This last aspect sets them apart from other non-parametric methods and explains their popularity in disciplines that deal with high-dimensional data, such as public health \cite{loef_using_2022}, labour economics \cite{hastie_random_2009, van_den_berg_predicting_2023}, and policy evaluation \cite{athey_recursive_2016, mcguire_predicting_2020}. It is also one of the reasons why Hal Varian, (ex) Google chief Economist, singles out Random Forests in his ``Tricks for Econometrics" paper \cite{varian_big_2014}.\\

\begin{figure}
    \centering
    \includegraphics[width=0.32\textwidth]{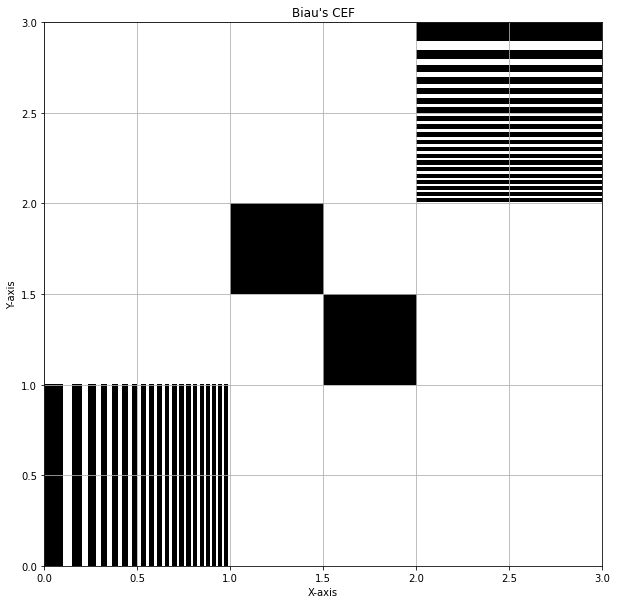}
    \includegraphics[width=0.32\textwidth]{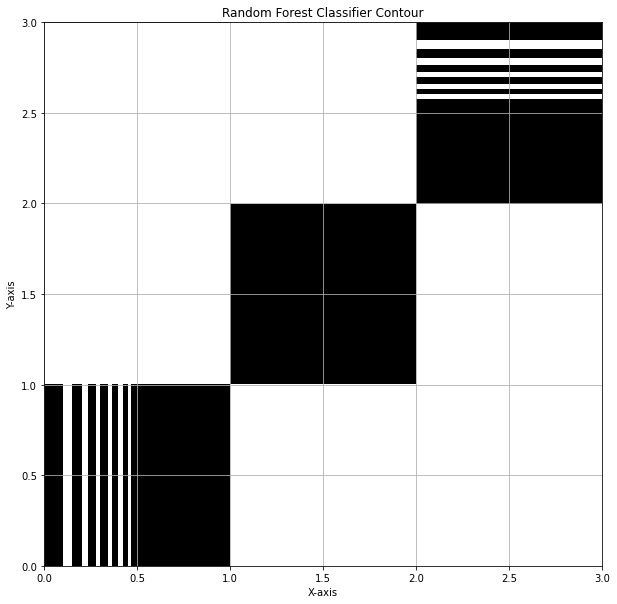}
    \includegraphics[width=0.32\textwidth]{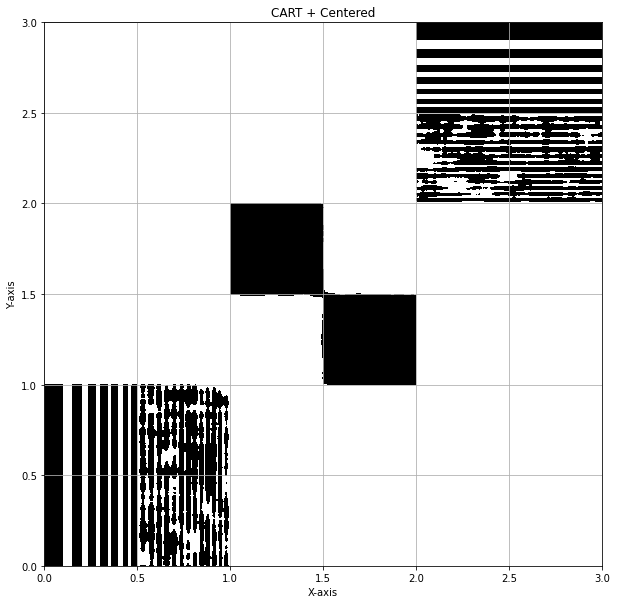}
    \caption{Contours of the Biau CEF, the Random Forest estimate, and the estimate of the variant studied in this paper. The bottom-left and top-right squares contain infinitely many stripes of decreasing width, so the CART algorithm will never split the checkerboard pattern in the middle square.}
    \label{fig:biauCEF}
\end{figure}

Yet, despite its popularity, \textit{``little is known about the mathematical properties of the method"} \cite{biau_random_2016}, and a large body of literature has examined the algorithm or variants of it under a variety of assumptions \cite{klusowski_universal_2021, duroux_impact_2018, mourtada_minimax_2019}. This paper serves as a contribution to this literature. In particular, an open question is when and if the Random Forest is consistent. There exists one major result \cite{scornet_consistency_2015} on the consistency of CART-based Random Forests. This result, and all variations on it, exists exclusively for the case of additive models, and when the resampling mechanism of the Random Forest is replaced with subsampling. It works by showing consistency of the CART algorithm under strong regularity assumptions. Indeed, further exploration of this shows that Random Forests perform catastrophically for such models, as the CART has difficulty learning the additive model well \cite{hastie_random_2009} and approaches that explicitly exploit the additive nature of the distribution would be directly preferred (see \cite{tan_doubly_2019}). As of yet, the question, whether the Breiman Random Forest is universally consistent is an open one. We know, however, that under the configurations studied in \cite{scornet_consistency_2015} the CART-based random forest is \textit{not} universally consistent, which can be shown by means of a clever example \cite{biau_consistency_2008} (see Figure \ref{fig:biauCEF}). \\

By consistency, here, we refer to $L^2$ consistency for non-parametric estimators, namely, that for a given distribution, the estimator of the regression function $\hat{m}_n(x)$, which is constructed from the sample (in this case a Random Forest), converges to the conditional expectation function $m(x)=\ex{Y|X=x}$ in the sense that 
$$\ex{m(X)-\hat{m}_n(X)}^2\to0\text{.}$$
Using fairly standard notation, the square here applies to the term inside the expectation operator. Universal consistency then refers to the case, where there is such a consistency guarantee regardless of the distribution.\\

Consistency of the method is important, as it is vital for inference, especially in settings, where Random Forests are applied to extract causal relationships (e.g., for estimating treatment effects). For this reason, consistency is often subsumed in applications. Random Forests, even though the method is two decades old, are seeing an increased uptake in recent applications specifically due to their suitability for the relatively novel Double Machine Learning approach \cite{chernozhukov_doubledebiased_2018}, where in fact Random Forests are provided as \textit{the} example methodology, because of their fast computation time and suitability for high dimensions. \\

Simultaneously, there is a large literature on the short-comings of the method. The most significant short-coming, known since the inception of the CART tree learner \cite{breiman_classification_1984}, is the so-called ``end cut preference" of the algorithm that leads to large bias of the algorithm along the edges of the feature space \cite{ishwaran_effect_2015}. In particular, CARTs are unable to pick up on high-frequency periodic patterns, which is usually not a problem with other non-parametric methods. Also, inherent to the method is a difficulty with picking up smooth, continuous patterns, due to the fact that tree learners are piece-wise constant functions and thus comparable to, e.g., histogram estimators. \\

For these reasons, a large literature studies variants of the method that address these problems. Generally, these fall into one of two categories, (1) toy models used as a  metaphor for the CART-based algorithm and (2) variants with specific use cases. The most studied variants are the Centered Forest \cite{breiman_consistency_2004}, where splits are performed at the median or midpoint of a node along a feature chosen at random, the Purely Random Forest \cite{biau_analysis_2012}, where both the feature and split points are fully random, Mondrian Forests \cite{mourtada_minimax_2019}, where the tree learners are constructed using Mondrian partitions, and Kernel Random Forests \cite{scornet_learning_2015}, where the splitting is replaced with an asymmetric kernel function. This list goes on (e.g., \cite{biau_analysis_2012, biau_neural_2018, lin_random_2006, zhou_deep_2020}).\\

Two other variants of the algorithm that are noteworthy are the Iterative Random Forest \cite{basu_iterative_2018}, where the method is iterated on iteratively reweighted features, which allows for the discovery of high-dimensional patterns. And, while not a Random Forest directly, although a direct connection has been conjectured \cite{breiman_random_2001}, Boosting \cite{breiman_bias_1996}, where tree stumps are trained on residuals until convergence.\\

This paper explores the suitability of ``grafting" consistent estimators onto the leaf nodes of a shallow CART step, where a particular emphasis will be placed on grafting on Centered Trees (or ``scions", as it were). As will be seen, 

\begin{enumerate}
\setstretch{1}
    \item This comes with a consistency guarantee of the entire algorithm, 
    \item It outperforms Random Forests in some settings, and otherwise delivers on-par performance,
    \item It allows traditionally unsuitable estimators like Kernel Regression to adapt to high-dimensional settings.
\end{enumerate}

\textbf{Composition of the paper}: After some setup, the main theoretical results are presented in Section 2, including some discussion over variations of the algorithm. In Section 3, the algorithm is applied in an empirical setting. Section 4 presents various numerical experiments, including discussion of the parameters of the method. Proofs and derivations are presented at the end in an Appendix for readability.

\section{Theoretical Results}

\subsection{Tools, Assumptions and Notation}

Throughout this paper, we will assume a setting, in which we have some (unknown) probability measure space $([0,1]^p\times \mathbb{R},\mathcal{F},\mathbb{P})$, where $\mathbb{P}$ is defined by a distribution, such that 

\[Y=m(X)+\varepsilon\]
with $m(x) = \ex{Y|X=x}$, $\varepsilon\sim \text{i.i.d.}(0,\sigma^2)$, and $X\sim\mathcal{U}[0,1]^p$. From this, we draw an i.i.d. sample $\mathcal{D}_n :=\{(X_i,Y_i)\}_{i=1}^n$ of $n$ independent copies of $(X,Y)$.\\

 It should be noted that these assumptions are fairly standard and that we impose them for theoretical elegance. Also, it should be pointed out that under these assumptions, consistency of the Breiman Random Forest has \textit{not} been proved and thus cannot be assumed. \vspace{10pt}\\
Further, we assume, \\
\textbf{Assumption 1}: $m_j(x)=\ex{Y|X^{(j)}=x}$ \textit{is $L_j$-Lipschitz}. \\
\textbf{Assumption 2}: \textit{We allow $p_n\to\infty$, however, we impose sparsity, namely that $Y$ depends only in $S$ relevant features, such that $\ex{Y|X}=\ex{Y|X^s}$ where $X^s$ is $S$-dimensional and only contains the $S$ relevant features.}\\

Note that Assumption 2 implies that the projection of $m$ onto any irrelevant feature is $m_j(x)=\ex{Y|X^{(j)}=x}=\ex{Y}$. Further we assume a functional form $Y(X)$, where $Y(X_i)=Y_i$. Also, let $\exn{F(X)}=\frac{1}{n}\sum_i F(X_i)$. Lastly, we refer to the $j$th component of a vector $x$ using the notation $x^{(j)}$.

\subsection{Random Forests}

Given $\mathcal{D}_n$, a random forest is an ensemble of recursively grown partition estimators, so-called trees or tree learners, $t_n(x;\theta,\dn)$, where $\theta$ is the (random) configuration used to grow the tree, which usually contains information on resampling and the order in which splits are performed. Let $\mathbf{\Theta}_n^M := (\theta^1_n, \theta^2_n,\ldots,\theta^M_n)$ be a vector of $M$ random draws of the configuration vector. The random forest regressor is then 
\[
\hat{m}_{n,M}(x;\mathbf{\Theta}^M_n,\dn)=\frac{1}{M}\sum_{l=1}^M{t_n(x;\theta^l_n,\dn)}\text{.}
\]

$t_n(x;\theta_n,\dn)$ partitions the feature space into so-called leaf nodes, which are typically hyperrectangles. To each leaf node, $t_n$ assigns a constant value, which serves as the prediction at that leaf node: specifically, the prediction is the average of the samples that fall into the leaf node. Let $A(x;\theta_n,\dn)$ be the leaf node that $x$ falls into. Further, included in $\theta_n$ is usually a resampling step, where a new sample of size $a_n\le n$ is drawn from $\dn$. We account for this explicitly. Let $s_i(\theta_n,\dn)$ be the sampling function, which is an integer factor between $0$ and $a_n$ measuring how often $(X_i,Y_i)$ was sampled for the construction of a specific tree $t_n(x;\theta_n,\mathcal{D}_n)$. It satisfies $\sum_i s_i(\theta_n,\dn) =a_n$. For sampling without replacement, $s_i(\theta_n,\dn)\in\{0,1\}$, for sampling with replacement $s_i(\theta_n,\dn)\in\{0,1,\ldots,a_n\}$. Then, it can be shown that the random forest has a local averaging form
\[
\hat{m}_{n,M}(x;\mathbf{\Theta}^M_n,\dn)=\sum_i^n {W_i(x)Y_i}\text{,}
\]
where $W_i(x)=\frac{s_i \mathbf{1}\{X_i\in A(x)\}}{\sum_j s_j \mathbf{1}\{X_j\in A(x)\}}$.

For purposes of notational decluttering, where it is obvious, we will often omit indices and the explicit dependence on the data $\dn$ and randomisation vector $\theta_n$ throughout the paper.\\

In classification settings, where $Y_i\in\{0,1\}$, $\hat{m}_{n,M}(x)$ is then used to construct a classifier
{\setstretch{1}
\begin{equation*}
    \hat{g}_{n,M}(x) = 
    \begin{cases}
        1 & \text{if } \hat{m}_{n,M}(x)>\frac{1}{2}\text{,}\\
        0 & \text{if } \hat{m}_{n,M}(x)\le\frac{1}{2}\text{.}
    \end{cases}
\end{equation*}}

To show consistency of the classifier, it is sufficient to show consistency of $\hat{m}_{n,M}(x)$.\\

As pointed out by Breiman originally \cite{breiman_random_2001} and in \cite{scornet_asymptotics_2016}, random forest algorithms interpolate well. In particular, adding more trees improves performance and we can show by a strong law of large numbers argument that 

\[
\hat{m}_{n,\infty}(x)=\lim_{M\to\infty}{\hat{m}_{n,M}(x)}=\exc{\theta_n}{t_n(x;\theta_n,\dn)}\eqqcolon \hat{m}_n(x)\text{.}
\]
$\exc{\theta_n}{\cdot}$ here refers to the expectation over the distribution of the configuration vector $\theta_n$. $\hat{m}_n(x)$ is the so-called infinite forest. It admits the local averaging representation 
\[
\hat{m}_n(x)=\sum_i w_i(x) Y_i\text{,}
\]
where we write $w_i(x) = \exc{\theta_n}{W_i(x)}$. This is the starting point of our analysis.

\subsubsection{Tree algorithms}

We consider three tree algorithms in this paper.\vspace{10pt}\\ 
\textbf{Algorithm (A)}: CART \cite{breiman_random_2001}:
\begin{enumerate}
\setstretch{1}
 \item Draw $a_n \leq n$ samples from $\mathcal{D}_n$ with replacement.
 \item Set the first node $A=[0,1]^p$ (root node).
 \item Split each node $A$ into two daughter nodes $A_L$ and $A_R$. For this, choose a set $T$ of $\nu$ test-coordinates uniformly at random from $\{1,2,...,p\}$
 and split the node at $$(j^*,z^*)=\arg \max_{\{(j,z):j\in T, z \in Z^j(A,\dn)\}} {L_n(j,z; A)}$$
 Where $Z^j(A,\dn)$ is the set of midpoints between the samples in $A$ along feature $j$ and 
\begin{align*}
    L_n(j,z;A) & = \exn{(Y(X) - \bar{Y}_A)^2|X\in A} \\ &- \frac{N(A_L)}{N(A)} \exn{(Y(X) - \bar{Y}_{A_L})^2|X\in A_L} \\&- \frac{N(A_R)}{N(A)}  \exn{(Y(X) - \bar{Y}_{A_R})^2|X\in A_R}
\end{align*}
is the (im)purity gain. Here, $A_L=\{x\in A: x^{(j)}\le z\}$ and $A_R = A\backslash A_L$,\\ $N(A)=\sum_i \mathbf{1}\{X_i \in A\}$ and $\bar{Y}_A=\exn{Y(X)|X\in A}$.
\item Continue splitting the resulting daughter nodes until (1) all samples in a node have the same value, (2) there is only one sample in a node, or (3) as long as the resultant nodes contain more than $q_n$ samples,  whichever happens first.
\item The final prediction is the mean at each leaf node.
\end{enumerate}
 
This is the original algorithm due to Breiman. $q_n$ here corresponds to the \texttt{minpoints} variable and $\nu$ to the \texttt{mtry} variable in the original implementation of the algorithm. Note that in the original paper, $q_n=1$, i.e., we have fully grown trees. Some discussion of $L_n(j,z;A)$, the CART-criterion, is in order at this juncture. Note that inspection of the the CART criterion reveals an alternate form (see Appendix):
$$L_n(j,z;A)=\frac{N(A_L)N(A_R)}{N(A)^2}\left[\bar{Y}_{A_L}-\bar{Y}_{A_R}\right]^2\text{,}$$

which is often the preferred implementation, given it requires less computation. Note that the CART prediction is simply the average over the samples in each leaf node, i.e., $\bar{Y}_{A}$, where $A$ using $L_n$. This shows that the CART criterion explicitly optimises over the difference between the resulting predictions of the tree, subject to a reweighting that punishes splits close to the edge. \\

We know by a law of large numbers argument that this converges to its population version almost surely:
$$L_n(j,z;A)\to \lambda(A_L)\lambda(A_R)\left[\ex{Y|X\in A_L}-\ex{Y|X\in A_R}\right]^2\equiv L(j,z;A)\text{,}$$
where $\lambda$ is the Lebesgue measure. If $L(j,z;A)$ has a unique maximum $(j^*,z^*)$, by reference to empirical process theory we expect $(j_n^*,z_n^*)\to^p(j^*,z^*)$, where $(j_n^*,z_n^*)$ refers to the maximiser of $L_n(j,z;A)$. This is investigated further in \cite{banerjee_confidence_2007}. This form also reveals that the CART criterion performs splits in the direction that results in the least amount of variation in the leaf nodes. \vspace{10pt} \\
\textbf{Algorithm (B)}: Centered (median) trees \cite{breiman_consistency_2004}:
\begin{enumerate}
\setstretch{1}
 \item Draw $a_n \leq n$ samples from $\mathcal{D}_n$ without replacement.
 \item Set the first node $A=[0,1]^p$ (root node). 
 \item Split each node $A$ into two daughter nodes. For this, take the samples that fall into $A$ and compute the sample median in each feature $\{1,2,...,p\}$. Then, choose a feature uniformly at random (i.e., with probability $1/p$), and split the node at the sample median in this feature.
 \item Continue splitting the resultant nodes until (1) all samples in a node have the same value, (2) there is only one sample in a node, or (3) as long as there are $q_n$ or more samples in the resultant nodes,  whichever happens first.
 \item The prediction at each node is the mean of the node sample.
\end{enumerate}
This is the algorithm that has been introduced by Breiman in 2004 for its theoretical elegance, to study a setting, in which Random Forests are consistent. It is perhaps the most studied variant of the Random Forest \cite{scornet_random_2016}. Note how the bootstrap regime is replaced by subsampling in this case. This is a common feature in most theoretical work on random forests, even in the study of the CART-based Breiman Random Forests \cite{scornet_consistency_2015}, as, unlike bootstrapping, where the resampling changes the distribution, the subsampled distribution is the same as that of the original data.\\

In \cite{duroux_impact_2018} it is derived that when $m$ is Lipschitz, $X$ is uniform on the $p$-hypercube, and $Y\in[-D,D]$ almost surely for some $D$, the algorithm is consistent with rate $n^{-\frac{\log{(1-3/(4p))}}{\log{2}-\log{(1-3/(4p)})}}$, whenever trees are grown to depth $\frac{1}{\log{2}-\log{(1-3/(4p))}}\left(\log{n} + C_1\right)$ and $a_n > C_2 n^{\frac{\log{2}}{\log{2}-\log{(1-3/(4p))}}}$, for some constants $C_1$ and $C_2$. Here, $\log(\cdot)$ refers to the natural logarithm. Also note that here, the Centered Forest is parametrised via its depth $k_n$ rather than leaf sample size $q_n$. That is, we control how often we may split to get to a leaf node. For Centered Forests, these two notions are equivalent, as each split effectively halves the sample, so $q_n\approx 2^{-k_n}a_n$.\\

We now move on to the algorithm studied in this paper: \vspace{10pt} \\
\textbf{Algorithm (C)}: Grafted Trees:
\begin{enumerate}
\setstretch{1}
 \item Draw $a_n \leq n$ samples from without replacement.
 \item Grow a tree using the CART criterion, such that the minimum number of leaf samples is $\alpha_n q_n$, where $\alpha_n\ge1$.
 \item Grow a Centered Forest on the leaf nodes of he CART step as long as there are at least $q_n$ in the resultant nodes.
\end{enumerate}
\textbf{Algorithm (C*)}: Generalized Grafting:
\begin{enumerate}
\setstretch{1}

 \item Draw $a_n \leq n$ samples from $\mathcal{D}_n$ without replacement.
 \item Grow a tree using the CART criterion, such that the minimum number of leaf samples is $\alpha_n q_n$, , where $\alpha_n\ge1$.
 \item Train a non-parametric regressor $\hat{\mu}_n(x)$ on the leaf nodes of the CART step.
 
\end{enumerate}

Note that due to its recursive nature, Algorithm (C) has  the same computation time benefits as the algorithm using full CARTs, namely by using multithreading. In fact, we even expect faster computation, as the random splits can be generated once beforehand and then just accessed from memory.\\

(C) and (C*) are the focus and contribution of this paper and we will now turn to them. Given our tree learner, we are interested in showing $L^2(\p)$ consistency: $$\ex{\hat{m}_n(X)-m(X)}^2\to0\text{.}$$

\subsection{Grafting, theoretical results}

We first show a general bias-variance decomposition theorem for (C) and our assumptions.

\textbf{Theorem 1}: \textit{Under Assumption 1 the $L^2(\p)$ error of the random forest regressor defined by }Algorithm (C) \textit{has the bound}
\begin{equation}
    \ex{\hat{m}_n(X)-m(X)}^2\leq p \sum_j{L_j^2 \ex{\ell_j(A(X))^2}} + \tilde{C}\sigma^2 \frac{a_n/q_n}{n}\text{,}
\end{equation}
\textit{where $\ell_j(A(x))=\sup_{y,z\in A}|y^{(j)}-z^{(j)}|$ is the $j$th side-length of the leaf node $A(x)$ containing $x$.}\\

\textbf{Remark 1}: \textit{In fact, under Assumption 2, we get }
$$\ex{\hat{m}_n(X)-m(X)}^2\leq S \sum_{j\in\mathcal{S}}{L_j^2 \ex{\ell_j(A(X))^2}} + \tilde{C}\sigma^2 \frac{a_n/q_n}{n}\text{,}$$ 
\textit{as $L_j=0$ for non-relevant features.}\\

We note two things about this. Firstly, the second term, which is the training error term, i.e., how quickly we ``denoise" the sample decays with $n^{-1}$ rather than $a_n^{-1}$, which is the rate for a single tree. This is the consequence of subsampling multiple (infinite) trees. Also, note how equation (1) gives us a potential starting point for understanding how the Grafted Forest with Centered ``scions" is able to pick up on high-frequency patterns in the data better than Breiman Random Forests. Through the process of repeatedly halving the nodes, we have a guarantee that the final side-lengths of the hyperrectangular partition sets decrease in all features. This is a guarantee that we don't generally have with CART \cite{biau_consistency_2008}.

\subsubsection{Algorithm (C), Grafted Trees}
We now move on to deriving consistency for the Grafted Tree algorithm defined in (C).

\textbf{Theorem 2}: Consistency of Algorithm (C). \textit{Under Assumption 1, for $p$ fixed, and for any choice of $\alpha_n$, and $a_n/q_n$, we have a worst bound on the $L^2(\p)$ error of}
\[
\ex{\hat{m}_n(X)-m(X)}^2\leq C_1p^2\left(1-\frac{5}{8p}\right)^{\log_2 {\alpha_n}}+C_2\frac{a_n/q_n}{n}\text{,}
\]
\textit{where $C_1$ and $C_2$ are independent of $p$ and $n$.}

A suitable choice of the parameters $a_n$, $q_n$ and $\alpha_n$ consistency then requires $\frac{a_n/q_n}{n}\to0$ and $\alpha_n\to \infty$. E.g.\ define $\alpha_n=2^{\kappa_n}$ and $q_n = a_n/2^{2\kappa_n}$, where $\kappa_n = c/2 \log_2 a_n$ for $c<1$. Then provided that $a_n^c/n\to0$, this yields consistency, which will hold for any legal choice of increasing $a_n$ with $a_n\le n$.

Note that the rate achieved by this algorithm, albeit a loose bound, beats the best rate we know for CART-based Random Forests \cite{klusowski_universal_2021}, which is of order $\log{n}^{-1}$.

\subsubsection{Algorithm (C*)}

\textbf{Theorem 3}: \textit{Let $\hat{\mu}_n$ be consistent with some rate $r(n_K)\to 0$ over all compact subsets $K$ of $[0,1]^p$, where $n_K$ is the number of samples in $K$, such that}
$$\ex{(\hat{\mu}_n(X)-m(X))^2|X\in K}\le r(n_K)\text{.}$$ \textit{Then,} Algorithm C* \textit{is consistent under our assumptions if $\alpha_nq_n \to \infty$.}\\

$r(n)$ here refers to the convergence rate of our estimator of choice, e.g.\ for a Nadaraya-Watson estimator with bandwidth $h_n\asymp n^{-\frac{1}{4+\beta}}$ for $\beta >0$, under our assumptions, we would have $r(n) = n^{-\frac{\beta}{4+\beta}}$ \cite{silverman_nonparametric_1993}. A note here is that we should expect the rate of convergence $r(n)$ to decay faster on the leaf nodes in general, given that the CART step systematically minimises the node variance, which we in turn expect to lead to a quicker convergence in the training error on the leaf node estimators.

\subsubsection{Sparse setting}
Another interesting question regards the sparsity setting implied by Assumption 2. The initial CART step serves both as a detector of discontinuity and a variance (or variation) reduction tool. We know that it can also serve as a feature selection tool. Indeed, under the sparsity setting in Assumption 2, it can be shown under some additional smoothness assumptions \cite{scornet_consistency_2015}, that for arbitrary $\xi$, $k$, and some corresponding $N_\xi$, whenever the sample size $n>N_\xi$, we know that with probability $1-\xi$, for the splits $j^*_l$ with $l<k$ we have 
$$j^*_{l}\in\mathcal{S}\text{.}$$
One direct way of using this, would be to train the grafted estimators only on the features selected in the CART step. If the CART selects the correct features, we know by Remark 1, that this would imply stable convergence of the algorithm \textit{independent} from $p_n$ which may increase arbitrarily quickly. This is explored in Section 4.

\section{Empirical Application}

In this section, the Grafted Tree algorithm is applied to the Boston Housing dataset, which is a widely used dataset in machine learning and statistics. It's a relatively small dataset containing information about homes in  the Boston suburbs. Each entry in the dataset corresponds to aggregated data about the $n=506$ neighborhoods or housing zones in the Boston area. For each suburb, 13 features, both numeric and categorical, and the outcome, the median value of owner-occupied homes, are documented. In more detail, the features are:

\begin{enumerate}
\setstretch{1}
    \item \texttt{CRIM}: per capita crime rate by town.
    \item \texttt{ZN}: proportion of residential land zoned for lots over 25,000 sq. ft.
    \item \texttt{INDUS}: proportion of non-retail business acres per town.
    \item \texttt{CHAS}: Charles River dummy variable (= 1 if tract bounds river; 0 otherwise).
    \item \texttt{NOX}: nitrogen oxide concentration (parts per 10 million).
    \item \texttt{RM}: average number of rooms per dwelling.
    \item \texttt{AGE}: proportion of owner-occupied units built prior to 1940.
    \item \texttt{DIS}: weighted distances to five Boston employment centers.
    \item \texttt{RAD}: index of accessibility to radial highways.
    \item \texttt{TAX}: full-value property tax rate per \$10,000.
    \item \texttt{PTRATIO}: pupil-teacher ratio by town.
    \item \texttt{B}: $1000(Bk - 0.63)^2$ where $Bk$ is the proportion of people of African American descent by town.
    \item \texttt{LSTAT}: \% lower status of the population.
\end{enumerate}

These features are then used to predict the outcome \texttt{MEDV}, which is the median value of owner-occupied homes in \$1000s. \\

For training, the dataset was split into a training set of 404 samples (roughly 80\% of the sample) and test set of 102 samples. We will refer to the indices of the test set as $I_{test}$ and correspondingly, the indices of the training set as $I_{train}$.\\

For all three algorithms a bootstrap sample size of $a_n=400$ was chosen and $M=100$ trees were grown. Three models were considered, the Breiman Random Forest (Algorithm (A)), and a bagged version, i.e.\ sampling with replacement of the Centered Forest (Algorithm (B)) and the Grafted Trees (Algorithm (C)). The parameters $q_n$ and $\alpha_n$ (for the Grafted Trees) were determined via 50-fold cross-validation, using \texttt{scikit}'s \texttt{RandomSearchCV} with the maximum amount of iterations. For the Breiman Random Forest $\nu$ was set to 13, i.e.\ the full feature set, based on the recommendation in \cite{genuer_variable_2010}. The result of this was an optimal choice of $q_n=1$ for the Breiman Random Forest (which corresponds to the original method in \cite{breiman_random_2001}), $q_n=3$ for the Centered Forest and $q_n=5$, $\alpha_n=4$ for the Grafted Trees.\\

Lastly, using the cross-validated parameters, the models were trained and the test error

$$Err_{test}(\hat{m}_{n,M},m)=\frac{1}{| I_{test}|}\sum_{i\in I_{test}} \left(Y_i - \hat{m}_{n,M}(X_i)\right)^2$$ was computed. The computed test error was $10.46$ for the Breiman Random Forest, $11.23$ for the Grafted Tree algorithm and $26.45$ for the Centered Forest algorithm. From this, we can clearly see that the Grafted Trees estimator vastly outperforms the Centered Forest, which is a widely-studied version of the Random Forest which due to its consistency guarantee. Also, the performance appears to be on-par with the Breiman Random Forest, albeit slightly worse in the specific test set.

\section{Experiments}
To gain a better understanding of the performance of the algorithm, we now turn to simulated data. For a conditional expectation function $m(X)$ of choice, we will sample $n$ data points $(X_i,Y_i)$ from a model 

$$Y = m(X) + \varepsilon\text{,}$$

where $\varepsilon\sim N(0,1)$ and $X\sim\mathcal{U}[0,1]^p$.

For analysis, the $L^2$ error is then computed over a mesh $\mathcal{X}_{mesh}$ of $[0,1]^p$ using 
$$Err(\hat{m}_{n,M},m)=\frac{1}{|\mathcal{X}_{mesh}|}\sum_{x\in \mathcal{X}_{mesh}} (\hat{m}_{n,M}(x)-m(x))^2\text{,}$$

where $\hat{m}_{n,M}$ is the estimator of interest. 

\subsection{\texorpdfstring{$L^2$}{} error as a function of \texorpdfstring{$n$}{}}
First, we wish to gain an understanding of the performance of the estimator defined by Algorithm (C) as the sample size increases. To this end, samples of increasing size were generated for different choices of $m(x)$, and $p=3$ was fixed. The algorithm was trained using $M=100$ trees and $a_n=\lceil n/1.3 \rceil$. $q_n$ and $\alpha_n$ were chosen via 50-fold cross-validation. For comparison, the Breiman Random Forest was also trained and its corresponding $q_n$ was also chosen via cross-validation. Lastly, the $L^2$ errors were plotted, alongside the optimal choice of $\alpha_n$ and the respective choices for $q_n$ for the Grafted Trees and the Breiman Random Forest. Some of these experiments are presented now.\\

\begin{figure}[H]
    \centering
    \includegraphics[width=0.8\textwidth]{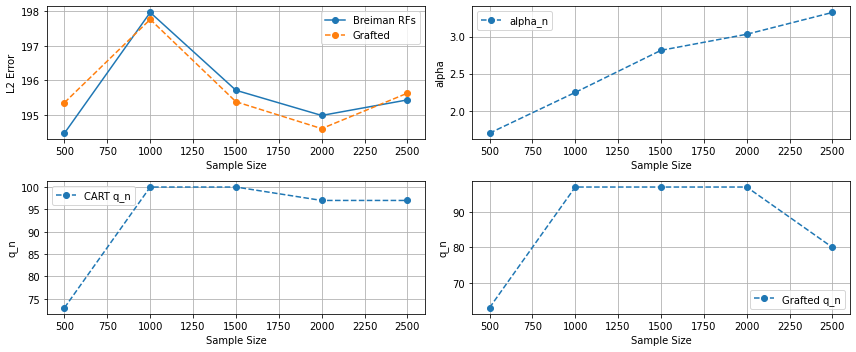}
    \caption{Plots for $m(x) = 100 \sin(200 x^{(1)}x^{(2)})$}
    \label{fig:din33}
\end{figure}

\begin{figure}[H]
    \centering
    \includegraphics[width=0.8\textwidth]{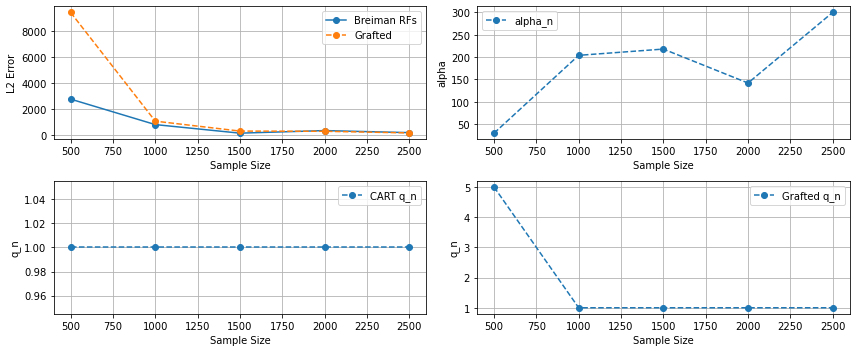}
    \caption{Plots for $m(x) = 100 (x^{(1)})^4$}
    \label{fig:100s}
\end{figure}

\begin{figure}[H]
    \centering
    \includegraphics[width=0.8\textwidth]{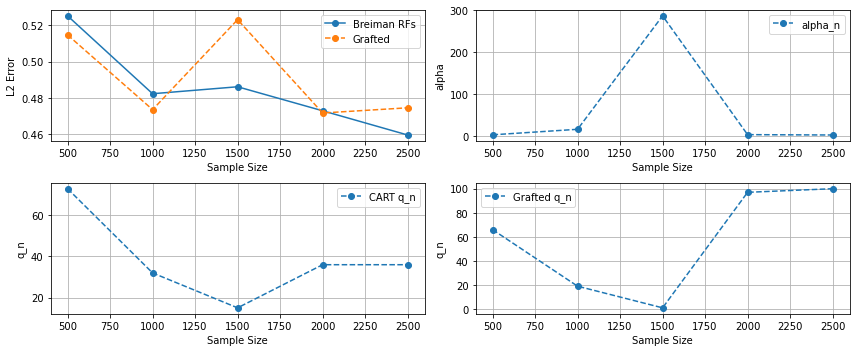}
    \caption{Plots for $m(x) = \cos(30(x^{(3)})^3)$}
    \label{fig:11}
\end{figure}

\begin{figure}[H]
    \centering
    \includegraphics[width=0.8\textwidth]{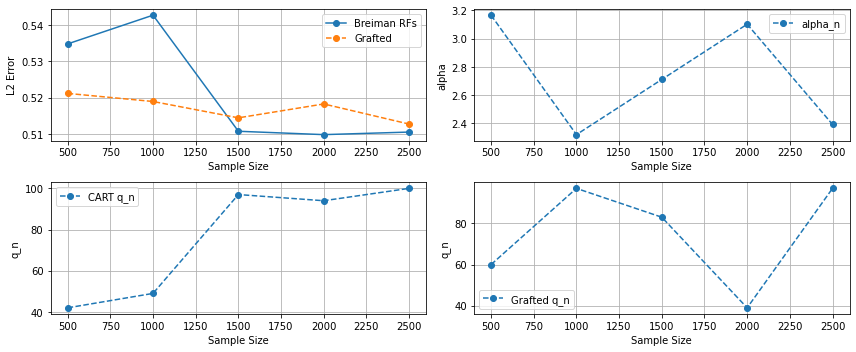}
    \caption{Plots for $m(x) = \cos(200x^{(1)} + x^{(2)})+x^{(3)}$}
    \label{fig:hff}
\end{figure}

\begin{figure}[H]
    \centering
    \includegraphics[width=0.8\textwidth]{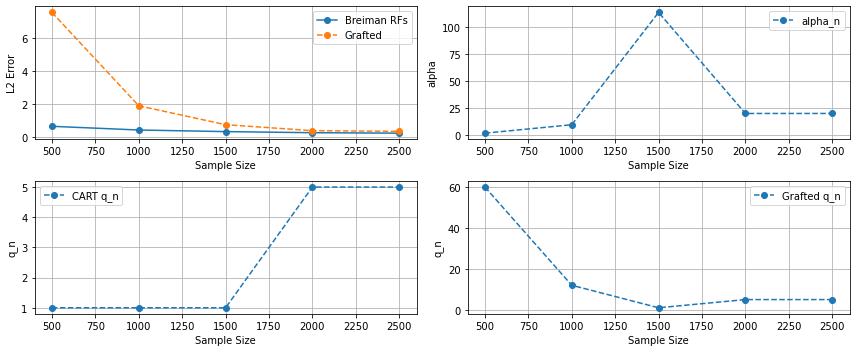}
    \caption{Plots for $m(x) = x^{(1)}x^{(2)}x^{(3)}$}
    \label{fig:hf}
\end{figure}

\begin{figure}[H]
    \centering
    \includegraphics[width=0.8\textwidth]{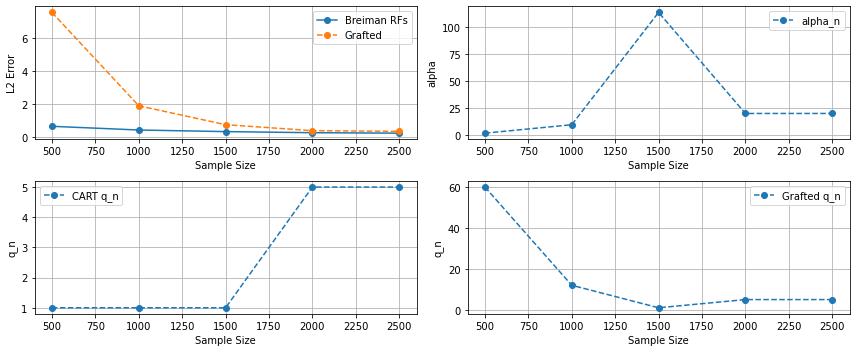}
    \caption{Plots for $m(x) = (x^{(1)})^2 +( x^{(2)} )^3$}
    \label{fig:hfsd}
\end{figure}

We find that the $L^2$ error decreases with sample size and that Grafted Trees demonstrate similar performance to Breiman Random Forests across the set of functions considered.

\subsection{Role of \texorpdfstring{$\alpha_n$}{}}

If we inspect the optimal value for $\alpha_n$ chosen in the cross-validation step of the experiments above, a common feature is that the optimal choice of $\alpha_n$ increases with sample size. This is promising, given that this is a necessary condition for consistency of the method. Further, we regularly encounter values of $\alpha_n$ above 10 or more, which seems to suggest that grafting is preferred over Breiman Random Forests in the scenarios studied. \\

\begin{figure}[H]
    \centering
    \includegraphics[width=0.8\textwidth]{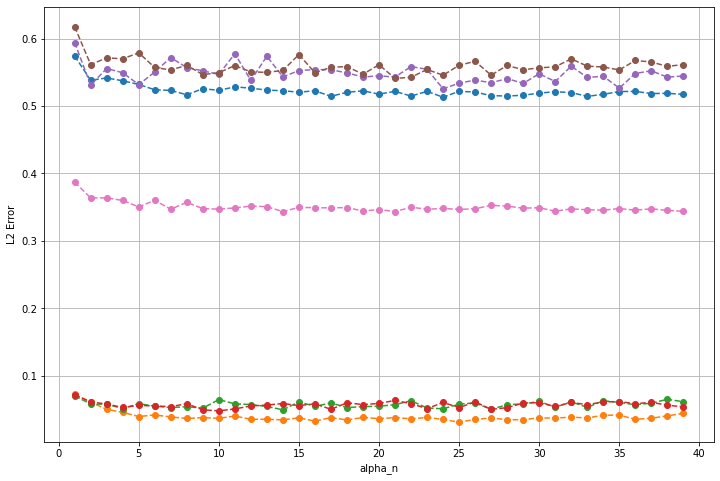}
    \caption{$L^2$ error for different values of $\alpha_n$, here, $q_n=10$. \vspace{4pt } The list of functions considered is $\sin(200 x^{(1)}x^{(2)})$, $x^{(1)}x^{(2)}$, $x^{(1)}x^{(2)} + x^{(3)}$, $\sin(x^{(1)}x^{(2)}) + \sin(x^{(3)})$, $-\cos((x^{(1)})^4(x^{(2)})^5x^{(3)}) + 0.2 (x^{(2)})^3$, and $\sin(x^{(1)}x^{(2)}) + x^{(1)}x^{(2)}x^{(3)}$.}
    \label{fig:cart_depth}
\end{figure}

To inspect this further, the $L^2$ error was examined for different values of $\alpha_n$ for seven functions of choice. For this, $q_n$ was set to $10$; $10,000$ samples were drawn and $100$ trees were grown with $a_n=8000$ and only $\alpha_n$ was varied. The results of this are presented in Figure \ref{fig:cart_depth}.\\

Looking at Figure \ref{fig:cart_depth}, it appears that in the settings studied, we prefer a shallower CART step as performance gets worse for deeper CART trees.

\subsection{Feature selection}
\begin{figure}[H]
    \centering
    \includegraphics[width=0.8\textwidth]{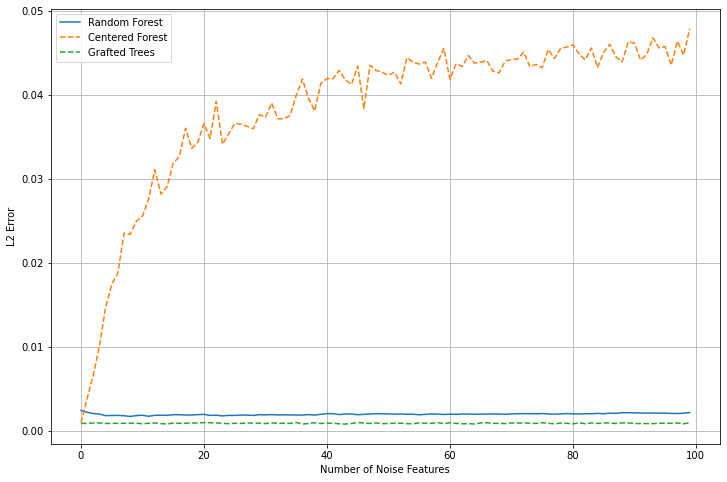}
    \caption{$L^2$ error as $p$ increases for the Grafted Trees method vs Centered Forests and Breiman Random Forests.}
    \label{fig:dimsCF}
\end{figure}

In this section, the viability of using the CART step to select the relevant features is examined. To this end, 1000 training samples were drawn from the distribution defined by 

$$Y=X^{(1)}X^{(2)}+\varepsilon$$
with $\varepsilon\sim N(0,0.1)$ and $X\sim\mathcal{U}[0,1]^p$, with $p\ge2$. Note that this reproduces the sparse setting in Assumption 2, namely with $S=2$. All values for $p$ ranging from $2$ to $102$ were considered, or equivalently, between $0$ and $100$ noise features were added. Then, the $L^2$ error was computed for Breiman's original Random Forests (i.e.\ $q_n=1$), Grafted Trees with $q_n=10$ and $\alpha_n=10$, and Centered Forests with $q_n=10$, all of which were trained on the same sample. For each, $M=100$ trees were grown and $a_n=\lceil n/1.3 \rceil$ was used. The results are plotted in Figure \ref{fig:dimsCF}.

We can clearly see that the $L^2$ error for both the Breiman Random Forest and the Grafted Trees is independent of the number of noise features. Indeed, the Grafted Trees outperform the Breiman Random Forest in this particular setup. We can also see that the Centered Forest $L^2$ error increases significantly with the number of redundant features. This confirms the usefulness of using the CART step for feature selection in Algorithm (C).

\subsubsection{Feature selection in Algorithm (C*)}
\begin{figure}[H]
    \centering
    \includegraphics[width=0.8\textwidth]{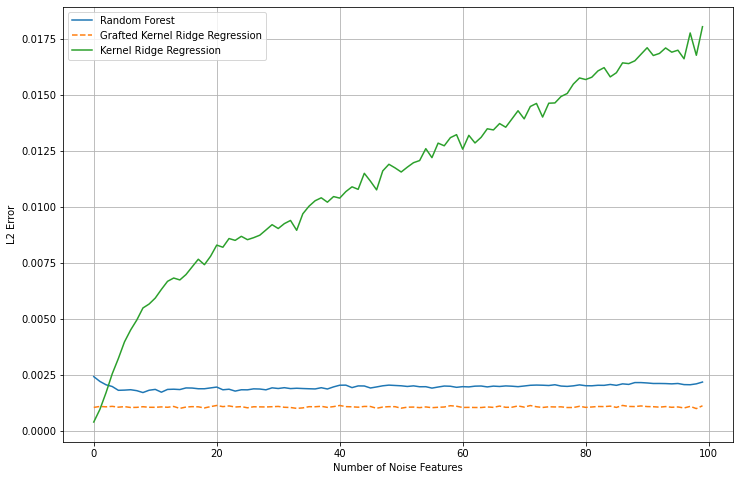}
    \caption{$L^2$ error as $p$ increases for the Grafted Kernel Ridge method vs plain Kernel Ridge Regression and Breiman Random Forests.}
    \label{fig:dimsNW}
\end{figure}

Similarly, we may interested in the viability of using feature selection for making consistent algorithms that are unsuitable for high-dimensional settings adapt to high dimensions by integrating them in an approach that uses Algorithm (C*). To explore this, the same setup as before was used but with Algorithm (C*) using a Kernel Ridge estimator on the leaf nodes, $\alpha_nq_n=100$, $M=100$ and $a_n=\lceil n/1.3 \rceil$. This was then compared to the $L^2$ performance of the plain Kernel Ridge estimator applied to the full dataset and Breiman's original Random Forest. The results are presented in Figure \ref{fig:dimsNW}.

We can clearly see that the grafted version of the Kernel Ridge estimator outperforms the plain version in the high dimensional setting considered and that its $L^2$ error is independent of the number of noise features added. Indeed, it also outperforms Breiman's original Random Forest in this setup.

\section{Conclusion and future work}

This paper explored the suitability of using Grafted Trees in prediction settings. Consistency was proved and the algorithm was empirically examined. A feature of this method that was not explored in this paper is the fact that in the limit, we expect tree construction to be independent from the data. Given that consistency has been shown, this makes the Grafted Trees an attractive candidate for extracting causal relatitionships, such as treatment effects, as this independence would obviate the need for, e.g., sample splitting.

\pagebreak

\bibliographystyle{plain}
\bibliography{main.bib}

\begin{thebibliography}{10}

\bibitem{athey_recursive_2016}
Susan Athey and Guido Imbens.
\newblock Recursive partitioning for heterogeneous causal effects.
\newblock {\em Proceedings of the National Academy of Sciences of the United States of America}, 113(27):7353--7360, July 2016.

\bibitem{banerjee_confidence_2007}
Moulinath Banerjee and Ian~W. McKeague.
\newblock Confidence sets for split points in decision trees.
\newblock {\em The Annals of Statistics}, 35(2), April 2007.
\newblock arXiv:0708.1820 [math, stat].

\bibitem{basu_iterative_2018}
Sumanta Basu, Karl Kumbier, James~B. Brown, and Bin Yu.
\newblock Iterative random forests to discover predictive and stable high-order interactions.
\newblock {\em Proceedings of the National Academy of Sciences}, 115(8):1943--1948, February 2018.
\newblock Publisher: Proceedings of the National Academy of Sciences.

\bibitem{biau_analysis_2012}
Gerard Biau.
\newblock Analysis of a {Random} {Forests} {Model}.
\newblock April 2012.

\bibitem{biau_consistency_2008}
Gérard Biau, Luc Devroye, and Gábor Lugosi.
\newblock Consistency of {Random} {Forests} and {Other} {Averaging} {Classifiers}.
\newblock {\em Journal of Machine Learning Research}, 9(66):2015--2033, 2008.

\bibitem{biau_random_2016}
Gérard Biau and Erwan Scornet.
\newblock A random forest guided tour.
\newblock {\em TEST}, 25(2):197--227, June 2016.

\bibitem{biau_neural_2018}
Gérard Biau, Erwan Scornet, and Johannes Welbl.
\newblock Neural {Random} {Forests}, April 2018.
\newblock arXiv:1604.07143 [cs, math, stat].

\bibitem{breiman_bias_1996}
Leo Breiman.
\newblock Bias, {Variance}, and {Arcing} {Classifiers}.
\newblock 1996.

\bibitem{breiman_random_2001}
Leo Breiman.
\newblock Random {Forests}.
\newblock {\em Machine Learning}, 45(1):5--32, October 2001.

\bibitem{breiman_consistency_2004}
Leo Breiman.
\newblock Consistency for a simple model of random forests.
\newblock 2004.
\newblock Publisher: Citeseer.

\bibitem{breiman_classification_1984}
Leo Breiman, Jerome Friedman, Charles~J. Stone, and R.~A. Olshen.
\newblock {\em Classification and {Regression} {Trees}}.
\newblock CRC Press LLC, Boca Raton, UNITED STATES, 1984.

\bibitem{chernozhukov_doubledebiased_2018}
Victor Chernozhukov, Denis Chetverikov, Mert Demirer, Esther Duflo, Christian Hansen, Whitney Newey, and James Robins.
\newblock Double/debiased machine learning for treatment and structural parameters.
\newblock {\em The Econometrics Journal}, 21(1):C1--C68, February 2018.

\bibitem{duroux_impact_2018}
Roxane Duroux and Erwan Scornet.
\newblock Impact of subsampling and tree depth on random forests.
\newblock {\em ESAIM: Probability and Statistics}, 22:96--128, 2018.
\newblock Publisher: EDP Sciences.

\bibitem{genuer_variable_2010}
Robin Genuer, Jean-Michel Poggi, and Christine Tuleau-Malot.
\newblock Variable selection using {Random} {Forests}.
\newblock {\em Pattern Recognition Letters}, 31(14):2225--2236, October 2010.
\newblock Publisher: Elsevier.

\bibitem{hastie_random_2009}
Trevor Hastie, Robert Tibshirani, and Jerome Friedman.
\newblock Random {Forests}.
\newblock In Trevor Hastie, Robert Tibshirani, and Jerome Friedman, editors, {\em The {Elements} of {Statistical} {Learning}: {Data} {Mining}, {Inference}, and {Prediction}}, Springer {Series} in {Statistics}, pages 587--604. Springer, New York, NY, 2009.

\bibitem{howard_two_2012}
Jeremy Howard and Mike Bowles.
\newblock The {Two} {Most} {Important} {Algorithms} in {Predictive} {Modeling} {Today}. {Strata} {Conference} {Presentation}, {New} {York}, {February} 28 2012. - {References} - {Scientific} {Research} {Publishing}, 2012.

\bibitem{ishwaran_effect_2015}
Hemant Ishwaran.
\newblock The effect of splitting on random forests.
\newblock {\em Machine Learning}, 99(1):75--118, April 2015.

\bibitem{klusowski_sparse_2020}
Jason Klusowski.
\newblock Sparse {Learning} with {CART}.
\newblock In {\em Advances in {Neural} {Information} {Processing} {Systems}}, volume~33, pages 11612--11622. Curran Associates, Inc., 2020.

\bibitem{klusowski_universal_2021}
Jason~M. Klusowski.
\newblock Universal {Consistency} of {Decision} {Trees} in {High} {Dimensions}, June 2021.
\newblock arXiv:2104.13881 [cs, math, stat].

\bibitem{lin_random_2006}
Yi~Lin and Yongho Jeon.
\newblock Random {Forests} and {Adaptive} {Nearest} {Neighbors}.
\newblock {\em Journal of the American Statistical Association}, 101(474):578--590, June 2006.
\newblock Publisher: Taylor \& Francis \_eprint: https://doi.org/10.1198/016214505000001230.

\bibitem{loef_using_2022}
Bette Loef, Albert Wong, Nicole A.~H. Janssen, Maciek Strak, Jurriaan Hoekstra, H.~Susan~J. Picavet, H.~C.~Hendriek Boshuizen, W.~M.~Monique Verschuren, and Gerrie-Cor~M. Herber.
\newblock Using random forest to identify longitudinal predictors of health in a 30-year cohort study.
\newblock {\em Scientific Reports}, 12(1):10372, June 2022.
\newblock Number: 1 Publisher: Nature Publishing Group.

\bibitem{mcguire_predicting_2020}
Shawn McGuire and Charles Delahunt.
\newblock Predicting {United} {States} {Policy} {Outcomes} with {Random} {Forests}, 2020.

\bibitem{mourtada_minimax_2019}
Jaouad Mourtada, Stéphane Gaïffas, and Erwan Scornet.
\newblock Minimax optimal rates for {Mondrian} trees and forests, April 2019.
\newblock arXiv:1803.05784 [math, stat].

\bibitem{scornet_learning_2015}
Erwan Scornet.
\newblock {\em Learning with random forests}.
\newblock PhD thesis, Université Pierre et Marie Curie - Paris VI, November 2015.

\bibitem{scornet_asymptotics_2016}
Erwan Scornet.
\newblock On the asymptotics of random forests.
\newblock {\em Journal of Multivariate Analysis}, 146:72--83, April 2016.

\bibitem{scornet_random_2016}
Erwan Scornet.
\newblock Random {Forests} and {Kernel} {Methods}.
\newblock {\em IEEE Transactions on Information Theory}, 62(3):1485--1500, March 2016.
\newblock Conference Name: IEEE Transactions on Information Theory.

\bibitem{scornet_consistency_2015}
Erwan Scornet, Gérard Biau, and Jean-Philippe Vert.
\newblock Consistency of random forests.
\newblock {\em The Annals of Statistics}, 43(4):1716--1741, August 2015.
\newblock Publisher: Institute of Mathematical Statistics.

\bibitem{silverman_nonparametric_1993}
P.~J.~Green Silverman, Bernard~W.
\newblock {\em Nonparametric {Regression} and {Generalized} {Linear} {Models}: {A} roughness penalty approach}.
\newblock Chapman and Hall/CRC, New York, May 1993.

\bibitem{tan_doubly_2019}
Zhiqiang Tan and Cun-Hui Zhang.
\newblock Doubly penalized estimation in additive regression with high-dimensional data.
\newblock {\em The Annals of Statistics}, 47(5):2567--2600, October 2019.
\newblock Publisher: Institute of Mathematical Statistics.

\bibitem{van_den_berg_predicting_2023}
Gerard~J. Van Den~Berg, Max Kunaschk, Julia Lang, Gesine Stephan, and Arne Uhlendorff.
\newblock Predicting {Re}-{Employment}: {Machine} {Learning} {Versus} {Assessments} by {Unemployed} {Workers} and by {Their} {Caseworkers}.
\newblock {\em SSRN Electronic Journal}, 2023.

\bibitem{varian_big_2014}
Hal~R. Varian.
\newblock Big {Data}: {New} {Tricks} for {Econometrics}.
\newblock {\em Journal of Economic Perspectives}, 28(2):3--28, May 2014.

\bibitem{zhou_deep_2020}
Zhi-Hua Zhou and Ji~Feng.
\newblock Deep {Forest}, July 2020.
\newblock arXiv:1702.08835 [cs, stat].

\end{thebibliography}

\appendix

\section*{Appendix}

\subsection*{Alternate form of the CART criterion}
{\allowdisplaybreaks\begin{align*}
L_n(j,z;A) &= \exn{(Y(X) - \bar{Y}_A)^2|X\in A} - \frac{N(A_L)}{N(A)} \exn{(Y(X) - \bar{Y}_{A_L})^2|X\in A_L} \\ &\:\:\:- \frac{N(A_R)}{N(A)}  \exn{(Y(X) - \bar{Y}_{A_R})^2|X\in A_R}\\
&=\exn{Y(X)^2|X\in A}-\bar{Y}_A^2- \frac{N(A_L)}{N(A)}\left(\exn{Y(X)^2|X\in A_L}-\bar{Y}_{A_L}^2\right)\\
&\:\:\:- \frac{N(A_R)}{N(A)}\left(\exn{Y(X)^2|X\in A_R}-\bar{Y}_{A_R}^2\right)\\
&=\exn{Y(X)^2|X\in A}-\left(\frac{N(A_L)}{N(A)}\bar{Y}_{A_L}+\frac{N(A_R)}{N(A)}\bar{Y}_{A_R}\right)^2\\
&\:\:\:- \frac{N(A_L)}{N(A)}\left(\exn{Y(X)^2|X\in A_L}-\bar{Y}_{A_L}^2\right)\\
&\:\:\:- \frac{N(A_R)}{N(A)}\left(\exn{Y(X)^2|X\in A_R}-\bar{Y}_{A_R}^2\right)\\
&=-\left(\frac{N(A_L)}{N(A)}\bar{Y}_{A_L}+\frac{N(A_R)}{N(A)}\bar{Y}_{A_R}\right)^2+\frac{N(A_L)}{N(A)}\bar{Y}_{A_L}^2+\frac{N(A_R)}{N(A)}\bar{Y}_{A_R}^2\\
&=\left(1-\frac{N(A_L)}{N(A)}\right)\frac{N(A_L)}{N(A)}\bar{Y}_{A_L}^2-2\frac{N(A_L)}{N(A)}\frac{N(A_R)}{N(A)}\bar{Y}_{A_L}\bar{Y}_{A_R}+\left(1-\frac{N(A_R)}{N(A)}\right)\frac{N(A_R)}{N(A)}\bar{Y}_{A_R}^2\\
&=\frac{N(A_L)}{N(A)}\frac{N(A_R)}{N(A)}\left(\bar{Y}_{A_L}-\bar{Y}_{A_R}\right)^2\text{.}
\end{align*}}

Using 
$$\exn{F(X)|X\in A}=\frac{N(A_L)}{N(A)}\exn{F(X)|X\in A_L} + \frac{N(A_R)}{N(A)}\exn{F(X)|X\in A_R}$$
and 
$$N(A)=N(A_L)+N(A_R)\text{.}$$

\subsection*{Proofs}

\textbf{Theorem 1}: We follow a classic bias-variance decomposition, as is usual in non-parametric theory. 
Define $\bar{m}_n(x)=\ex{\hat{m}_n(x)|X_1,X_2,...,X_n}$, i.e.\ the ``denoised" version of the estimator. In particular, note here that 
$$\bar{m}_n(x)=\sum_i w_i(x) m(X_i)$$

Then note that for the $L^2(\p)$ error we have
$$\ex{\hat{m}_n(X)-m(X)}^2=\ex{m(X)-\bar{m}_n(X)}^2+\ex{\hat{m}_n(X)-\bar{m}_n(X)}^2$$

The first part we may refer to as the estimation error, the second as the training error. We know from intuitive reasoning, that the training error should decay with $\frac{\sigma^2}{q_n}$. We now consider these two components in turn:\\

(1) Estimation error:

$$\ex{m(X)-\bar{m}_n(X)}^2=\ex{\exc{\theta}{\sum_i W_i(X) [m(X)-m(X_i)]}}^2$$

Consider $\sum_i W_i(x) [m(X)-m(X_i)]$ assuming that we are given a realisation of $\theta$. We have
\begin{align*}
\sum_i W_i(X) [m(X)-m(X_i)]&=\sum_{i\in I(A(X))} W_i(X) [m(X)-m(X_i)]\\
&\le\sum_{i\in I(A(X))} W_i(X) \sum_{j=1}^p {L_j |X^{(j)}_i-X^{(j)}|}\\
&\le\sum_{i\in I(A(X))} W_i(X) \sum_{j=1}^p {L_j \sup_{x,z\in A(X)}{|y^{(j)}-z^{(j)}|}}\\
&=\sum_{j=1}^p {L_j \ell_j(A(X))}\text{.}
\end{align*}

Here, $I(A(X))$ is a set of indices for the samples falling into $A(X)$. Then, note that

\begin{align*}
\ex{m(X)-\bar{m}_n(X)}^2&=\ex{\exc{\theta}{\sum_i W_i(X) [m(X)-m(X_i)]}}^2\\
&\le\ex{\exc{\theta}{\sum_{j=1}^p {L_j \ell_j(A(X))}}}^2\\
&\le p\sum_{j=1}^p L_j^2\ex{{\ell_j(A(X))}}^2\\
&\le p\sum_{j=1}^p L_j^2\ex{\ell_j(A(X))^2}\text{,}
\end{align*}

using the Cauchy-Schwarz inequality and Jensen's inequality. Note that in $\ex{\ell_j(A(X))^2}$ we take an expectation over all randomness, the partitioning and \textit{and} random sampling.\\

(2) Training error:

\begin{align*}
    \ex{\hat{m}_n(X)-\bar{m}_n(X)}^2&=\ex{\sum_i w_i(X)[Y_i-m(X_i)]}^2\\
    &=\ex{\sum_i w_i(X)[\varepsilon_i]}^2\\
    &=\ex{\sum_{i\in I(A(X))} w_i(X)\varepsilon_i}^2\\
    &\le\ex{\sum_{i\in I(A(X))} w_i(X)^2\varepsilon_i^2}\\
    &\le\sigma^2\ex{\sum_{i\in I(A(X))} w_i(X)^2}\\
    &\le\sigma^2\ex{\sum_{i\in I(A(X))} w_i(X)\max_iw_i(X)}\\
    &\le\sigma^2\ex{\max_iw_i(X)}\text{.}
\end{align*}

Now by definition, $w_i(X)=\exc{\theta}{\frac{s_i \mathbf{1}\{X_i\in A(X)\}}{\sum_js_j \mathbf{1}\{X_j\in A(X)\}}}$. \\
We know $\sum_js_j \mathbf{1}\{X_j\in A(X)\}\ge q_n$ and $s_i \mathbf{1}\{X_i\in A(X)\}\le s_i$. Together, this gives $w_i(X) \le \exc{\theta}{s_i} / \lceil q_n\rceil$. Using this, then

\begin{align*}
\ex{\max_i w_i(X)} &\le \frac{1}{\lceil q_n \rceil} \max_i \exc{\theta}{s_i}\\
&\le\tilde{C}\times1/q_n \times a_n/n\text{.}
\end{align*}

For which we have used, that the sampling probability is $a_n/n$. Thus, we have proved the theorem. $\square$

\textbf{Theorem 2}: To prove the consistency of the Grafted Tree, we exploit regularity in the centered ``scion" splits. This is similar to \cite{duroux_impact_2018} with the significant difference that we need to account for the CART step and leaf sample size. 

Consider the successive recursive partition sets containing $X$ as we traverse the tree. They are hyperrectangles of decreasing side-length. Consider the distribution of the side length as we traverse the tree. Now, the first splits use the CART criterion. Let the intermediate side-length in the $j$th feature after the CART step terminates be distributed as $\mathcal{L}_{CART}^j$, which is some distribution that is a fully determined function of the data, $\theta$ and $\alpha_nq_n$. We know that $\mathcal{L}_{CART}^j\le1$.

After the CART splits, we have independent splits at the median of some randomly drawn feature. Say, we know exactly, that $n_k$ samples are in the $k$th node of the centered step containing $X$. In particular, $n_0\ge \alpha_nq_n$ is the leaf sample size. The node has some length $\mathcal{L}_{k-1}$ and the split is at the median, so if the split occurs again in this feature, the length is distributed as $\mathcal{L}_{k-1}\beta(n_k + 1, n_{k-1} - n_k)$, where $\beta(a,b)$ is the beta distribution. Using this, we know that 

$$\ell_j(A(X)) \sim \mathcal{L}_{CART}^j \times \Pi_{k>\kappa_n}\beta(n_k + 1, n_{k-1} - n_k)^{\mathbf{1}\{j^*_k = j\}}\text{,}$$

where $j^*_k$ is the feature selected for the $k$th split. Now, as $n_{k-1}/n_k \ge 2$ the expectation clearly vanishes:
\begin{align*}
   \ex{\ell_j(A(X))}&=\ex{\mathcal{L}_{CART}^j} \times \Pi_{k}\ex{\beta(n_k + 1, n_{k-1} - n_k)^{\mathbf{1}\{j^*_k = j\}}}\\
   &\le\Pi_{k}\left[\frac{p-1}{p}+\frac{1}{p}\ex{\beta(n_k + 1, n_{k-1} - n_k)}\right]\\
   &=\Pi_{k}\left[\frac{p-1}{p}+\frac{1}{p}\frac{n_k + 1}{n_{k-1} + 1}\right]\\
   &\le\Pi_{k}\left[\frac{p-1}{p}+\frac{1}{2p}\right]\xrightarrow[]{k\to\infty}0\text{.}
\end{align*}

We need
\begin{align*}
   \ex{\ell_j(A(X))^2}&=\ex{(\mathcal{L}_{CART}^j)^2} \times \Pi_{k}\ex{\beta(n_k + 1, n_{k-1} - n_k)^{2\times\mathbf{1}\{j^*_k = j\}}}\\
   &\le\Pi_{k}\left[\frac{p-1}{p}+\frac{1}{p}\ex{\beta(n_k + 1, n_{k-1} - n_k)^2}\right]\\
   &=\Pi_{k}\left[\frac{p-1}{p}+\frac{1}{p}\frac{(n_k + 1)(n_k + 2)}{(n_{k-1} + 1)(n_{k-1} + 2)}\right]\\
   &\le\Pi_{k}\left[\frac{p-1}{p}+\frac{1}{4p}\frac{n_{k-1} + 4}{n_{k-1} + 1}\right]\\
   &\le\Pi_{k}\left[\frac{p-1}{p}+\frac{5}{8p}\right]\text{.}
\end{align*}

Clearly, $\left[\frac{p-1}{p}+\frac{5}{8p}\right]<0$. Then, $\Pi_{k}\left[\frac{p-1}{p}+\frac{5}{8p}\right]\le\left[1-\frac{5}{8p}\right]^{\lfloor\log_2{\alpha_n}\rfloor}$. Together with Theorem 1, we then get the required result.

\textbf{Theorem 3}: Consider a single tree learner. Let $\mathcal{P}$ be the partition induced in the CART step, we have
$$\ex{t_n(X)-m(X)}^2=\ex{\ex{(t_n(X)-m(X))^2|\mathcal{P}}}\text{.}$$
This partition is completely determined by the data and $\theta$. Then,
\begin{align*}
    \ex{(t_n(X)-m(X))^2|\mathcal{P}}&=\sum_{A\in\mathcal{P}}\lambda(A)\ex{(\hat{\mu}_n(X;A)-m(X))^2|X\in A}\\
    &\le \sum_{A\in\mathcal{P}}\lambda(A)\ex{(\hat{\mu}_n(X;A)-m(X))^2|X\in \bar{A}}\\
    &\le r(\alpha_nq_n)\text{,}\\
\end{align*}
where $\hat{\mu}(x;A)$ is the estimator on the leaf sample of $A$, $\bar{A}$ is the closure of $A$, which is compact.

By extension, then, the average over the $M$ trees is also consistent.

\end{document}